\pdfoutput=1

\documentclass[11pt]{article}

\usepackage[final]{acl}

\usepackage{times}
\usepackage{latexsym}
\usepackage{xurl}
\usepackage{hyperref}
\usepackage{amsmath}
\usepackage{verbatim}
\usepackage[T1]{fontenc}

\usepackage[utf8]{inputenc}

\usepackage{microtype}

\usepackage{inconsolata}

\usepackage{graphicx}
\usepackage{svg}
\usepackage{cleveref}
\usepackage{booktabs}
\usepackage{multirow}
\usepackage{amsfonts}
\usepackage{subfigure}
\usepackage{enumitem}

\newcommand{\judgment}{$\mathsf{Judgment}$ }
\newcommand{\navigation}{$\mathsf{Navigation}$ }

\title{How Accurate Are LLMs at Multi-Question Answering \\ on Conversational Transcripts?}

\author{Xiliang Zhu, \ Shi Zong, \ David Rossouw \\
    Dialpad Inc. \\
    \texttt{\{xzhu,shi.zong,davidr\}@dialpad.com}}

\begin{document}

\maketitle

\begin{abstract}

Deploying Large Language Models (LLMs) for question answering (QA) over lengthy contexts is a significant challenge. In industrial settings, this process is often hindered by high computational costs and latency, especially when multiple questions must be answered based on the same context. In this work, we explore the capabilities of LLMs to answer multiple questions based on the same conversational context. We conduct extensive experiments and benchmark a range of both proprietary and public models on this challenging task. Our findings highlight that while strong proprietary LLMs like GPT-4o achieve the best overall performance, fine-tuned public LLMs with up to 8 billion parameters can surpass GPT-4o in accuracy, which demonstrates their potential for transparent and cost-effective deployment in real-world applications. 
\end{abstract}

\begin{figure*}[t]
\centering
\includegraphics[width=0.995\textwidth]{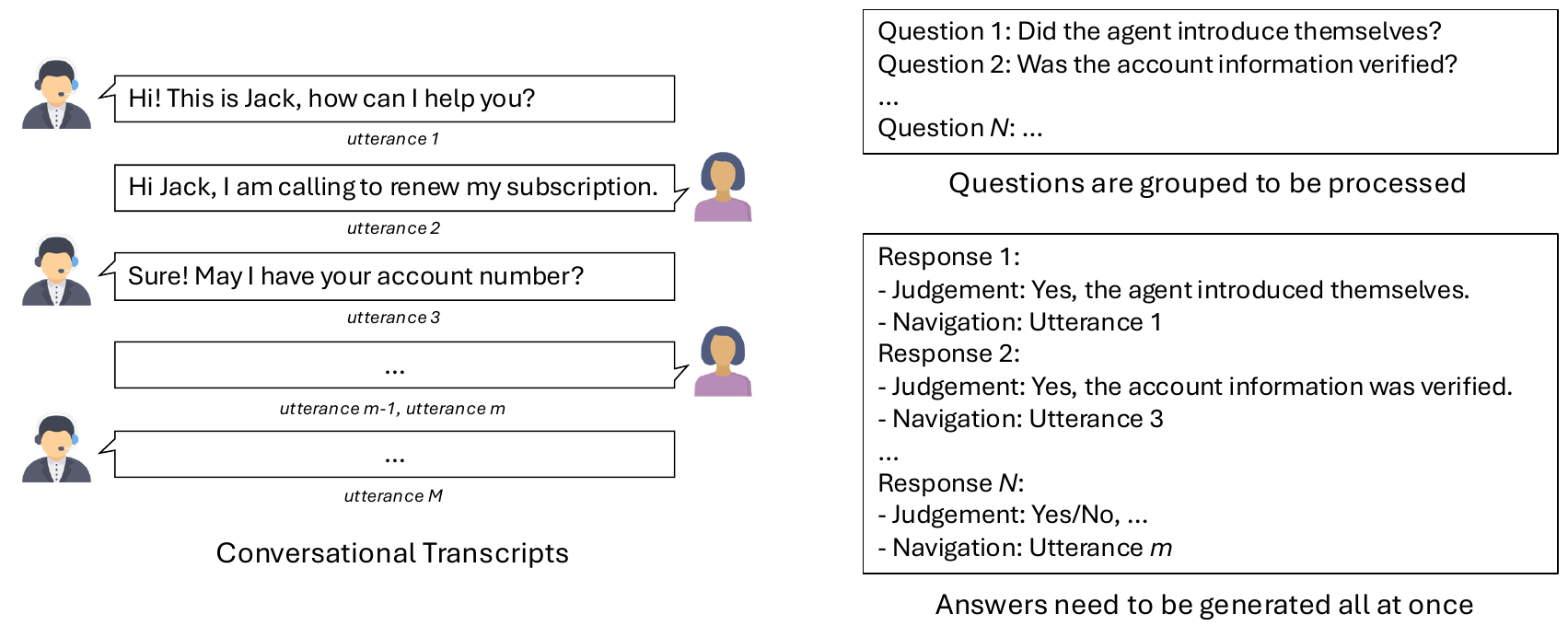}
\caption{Illustration of our task. The inputs of the model are conversational transcripts in contact centers and questions of group size $N$, and then we instruct the model to generate the responses to these questions in a single prompt. The model is instructed to generate each response by providing a ``yes'' or ``no'' judgment ($\mathsf{Judgment}$) and the supporting utterance from the original transcripts ($\mathsf{Navigation}$).}
\label{fig_overview}
\end{figure*}

\section{Introduction}
\label{sec:introduction}

Question Answering (QA) is a pivotal task in Natural Language Processing (NLP), which involves extraction or generation of responses from a given context in reply to user queries. It has been proven invaluable in a diverse array of applications across various industries, such as healthcare, finance, legal and customer support \cite{jin2020diseasedoespatienthave, li2024experimentinglegalaisolutions, Chen_2024, castelli-etal-2020-techqa}.

One real-world use case of QA in the customer support domain is producing responses to a set of questions about the agent-customer conversations. 
This allows for the identification of key components and topics within a conversation in the customer support scenario, helping locate the relevant discussion in the conversation efficiently. 
Such QA systems enable contact center supervisors to pinpoint potential coaching opportunities, improving agents' ability to better respond to customers' queries and navigate the conversation flow. 
For instance, verification of the customer's information is often a standard protocol in contact center calls. A question like ``\textit{Was the customer's account information verified?}'' helps contact center supervisors identify if such protocol is followed in the calls, thus necessary coaching and training can be provided. This approach not only improves operational efficiency but also improves the overall quality of customer service.

With the rapid development of Large Language Models (LLMs), there has been an increasing interest in utilizing LLMs in the QA task.
However, deploying LLMs in an industrial production environment presents significant challenges, primarily due to high inference costs and latency. 
In the context of our QA use case within the customer support domain, we often have multiple questions associated with a single, often lengthy, conversation transcript (over 25\% of transcripts contain more than 2,700 tokens; more details in \S\ref{subsec:dataset_development}). 
Running separate inferences for each question with the same long contextual background creates considerable overhead for the QA system. Therefore, optimizing the inference workflows without compromising the accuracy is crucial for real-world applications. 

One natural and intuitive idea to address the above challenge is to combine multiple questions within one single run.
Some prior work indeed takes this approach, for example, batch prompting introduced by \citet{cheng-etal-2023-batch, lin2024batchprompt}.
However, several research questions still remain open.
First, to the best of our knowledge, existing empirical evaluations have not considered realistic QA scenarios involving lengthy, multi-turn conversational contexts such as customer-support call-center transcripts. It thus remains unclear whether batch prompting effectively scales to industrial settings, where a single transcript often spans thousands of tokens and triggers dozens of questions.
Second, existing studies on batch prompting focus primarily on proprietary LLM APIs with limited in-context learning, leaving the potential of public and smaller models unexplored.

In this work, we focus on understanding the capability of LLMs for generating multiple structured responses to a set of questions within a single prompt, based on real-world conversational transcripts. 
Each response not only includes the answer to the question ($\mathsf{Judgment}$), but also a verifiable reference from the conversation context to enable human review and mitigate hallucination ($\mathsf{Navigation}$). 
We conduct extensive experiments with several proprietary and public LLMs, and benchmark their performance under both zero-shot and fine-tuned settings.

In summary, our contributions are as follows:
\begin{itemize}
    \item We adapt the recently-proposed batch prompting paradigm and explore its effectiveness on real-world long conversational transcripts.
    \item We conduct a comprehensive benchmark comparing leading proprietary LLMs (e.g. GPT-4o) against fine-tuned public LLMs (e.g. Llama-3) on this conversation-based QA task, assessing their ability to process up to 50 questions within a single prompt.
    \item Our extensive experiments demonstrate that fine-tuned public models with as few as 8 billion parameters can surpass the \judgment accuracy of GPT-4o, which reveals the remarkable potential of smaller public models.
\end{itemize}
We hope this study will be a useful guide for industry practitioners in making informed decisions regarding model selection and batch size configuration when deploying large‑scale QA pipelines for extended customer‑support dialogues.

\section{Related Work}
\label{sec:related_work}

There has been a long history of studying the QA task.
Based on the information source where answers are from, the QA task could be categorized into two types: textual QA and knowledge base QA \cite{zhu2021retrievingreadingcomprehensivesurvey}.

Over the years, some other variations have been proposed, including multi-hop QA \cite{yang-etal-2018-hotpotqa}, visual QA \cite{Antol_2015_ICCV}, and table QA \cite{mueller-etal-2019-answering}. 
In this section, we focus on making distinctions between our task and existing tasks in literature and refer readers to recent surveys (for example \citet{zhu2021retrievingreadingcomprehensivesurvey, 9960856}) for a more comprehensive review of different QA tasks.

Our task is based on conversational transcripts collected in contact centers. It is different from conversational QA in literature \cite{reddy-etal-2019-coqa, CQA}, which engages users in multiple rounds of questions and answers with the model.
Moreover, these transcripts are often lengthy and noisy as speech recognition errors are often present.
Due to the conversational nature of the transcripts, these contents are also less structured. All of these bring unique challenges compared to the long-context QA that normally uses documents as main resources \cite{wang-etal-2024-leave}.

In our product use case, we primarily focus on questions that can be directly answered by ``yes'' or ``no''. Although this type of question has been previously studied by Yes/No QA \cite{clark-etal-2019-boolq}, our main focus is on generating responses for all questions \textit{within one single run} under an industrial environment, instead of processing questions one by one.
Additionally, to provide further explainability of the generated binary judgment, we also instruct models to select supporting evidence from the transcripts (to be introduced in \S\ref{subsec:task_overview}).


\section{Our Task}

\subsection{Task Overview}
\label{subsec:task_overview}

Our task utilizes LLMs to analyze human conversational transcripts in contact centers. 
These call transcripts are generated by our in-house Automatic Speech Recognition (ASR) engine. 
Contact center supervisors can then set up a collection of $N$ questions (defined as \textit{group size $N$}) about the conversations in the contact center and for each question, a ``yes'' or ``no'' judgment is expected (\judgment in the response). 
Our current setup of considering only binary Yes/No responses is based on the feedback from our clients, thus reflecting the actual demand from the market.
We provide some example questions in \Cref{sec:appendix_example_questions}.

To enhance the explainability and verifiability of each response and facilitate human review for potential hallucinations, we further introduce an additional \navigation component in the generated responses. 
Specifically, a transcript is segmented into $M$ utterances,\footnote{An utterance starts when speech begins and ends when a significant pause occurs. Speech from one single person could be split into several utterances and our ASR engine automatically determines where to make these splits.} each of which is assigned an index starting from 1 ($m \in \{1, 2, ..., M\}$). 
Beyond the binary judgment (``yes'' and ``no'') discussed above, LLMs are also instructed to provide one index of the utterance that best supports this judgment. 
An overview of our task is in \Cref{fig_overview}.

\subsection{Structured Output}
\label{subsec:our_method}

In our pipeline, we directly prompt LLMs: $N$ questions and the transcript are provided as the context in the prompt, and models are tasked with generating the corresponding $N$ responses for each of the questions.

As multiple answers are retrieved in a single LLM inference pass, in order to generate robust responses in an industrial setting, we develop a JSON-based response structure to include all requested information (\judgment and \navigation), and is reinforced by an in-context formatting example that instructs the LLMs to generate such structure reliably.
The template of our designed prompt and the expected response structure are provided in \Cref{sec:appendix_A}.


\section{Experiments}

\subsection{Dataset Development}
\label{subsec:dataset_development}

\paragraph{Data Collection.} 

We collect a total of 2,800 conversation transcripts produced by our ASR engine from various contact centers. To better understand the distribution of transcript lengths from these calls, we calculate the number of tokens for different percentiles in \Cref{tab:internl_dataset_stats}. 
We notice that more than 25\% of our transcripts exceed 2,700 tokens, with notably 5\% of the total transcripts extending to as many as 6,500 tokens.

\begin{table}
\centering
\small
\begin{tabular}{c|r}
\toprule
Percentile          & \# of Tokens  \\
\midrule
25\textsuperscript{th} & 501 \\ 
50\textsuperscript{th} & 1,153 \\ 
75\textsuperscript{th} & 2,754 \\ 
95\textsuperscript{th} & 6,584 \\ 
\bottomrule
\end{tabular}
\caption{Statistics on number of tokens of our internal transcript data, with \textit{tiktoken}\protect\footnotemark{} used as the tokenizer.}
\label{tab:internl_dataset_stats}
\end{table}
\footnotetext{The tiktoken tokenizer can be accessed at: \url{https://github.com/openai/tiktoken}}

\begin{table}
\centering
\small
\begin{tabular}{c|c|c}
\toprule
group size $N$ & Judgment Acc & Navigation Acc \\
\midrule
5      & 0.96  & 0.90  \\
10     & 0.95  & 0.91   \\
20     & 0.91  & 0.86   \\
30     & 0.88  & 0.85   \\
40     & 0.84  & 0.81   \\
50     & 0.84  & 0.75   \\
\bottomrule
\end{tabular}
\caption{Human evaluation on Judgment and Navigation accuracy of GPT-4o, with varying group size $N$. An inter-annotator agreement of 0.85 is achieved for this annotation.}
\label{tab:gpt-4o human review}
\end{table}

\paragraph{Annotation.}

Given the long-context nature of our transcript that discussed above, it is impractical to conduct large-scale human annotations for generating answers to all the questions paired with these transcripts.

Since GPT-4o \cite{openai2024gpt4ocard} is broadly recognized as one of the top-performing LLMs across various NLP tasks, to address the annotation challenges, we did an initial round of human evaluation on GPT-4o responses to check whether GPT-4o is a strong and reliable baseline for this task. We present the human evaluation on GPT-4o responses over 50 call transcripts in \Cref{tab:gpt-4o human review}. We observe that GPT-4o excels in accuracy with a group size of 5 or 10, both achieving high accuracy of $\sim$0.95 and $\sim$0.90 for $\mathsf{Judgment}$ and $\mathsf{Navigation}$ respectively. However, it presents noticeable decreasing performance for both $\mathsf{Judgment}$ and $\mathsf{Navigation}$ when the group size becomes larger.
Thus, in this study, we group all questions into sets of 10 and use the resulting responses generated by GPT-4o as the standard reference for training and evaluating other models. In other words, we divide 50 questions that are paired with each transcript into five folds, resulting in five rounds of API calls (each containing 10 different questions). In this way, we are able to achieve a balance between maintaining annotation accuracy and reducing the LLM API cost.

\paragraph{Train/dev/test Splits.} 

We sample 2,000, 500 and 300 call transcripts to serve as our training, testing and development sets, respectively.

The goal of this study is to analyze the quality of the generated responses from different models, with different numbers of questions grouped in the prompt (i.e. group size $N$ defined in \S\ref{subsec:task_overview}).
Thus, when compiling the test and development sets, for each group size \( N \in \{10, 20, 30, 40, 50\} \), we randomly select $N$ question-answer pairs to pair with each transcript (i.e., there are $300\times 10=3,000$ questions in the test set when the group size $N$ is set to 10).
This enables a granular evaluation of model performance when the group size varies in the prompt.

\begin{figure*}[h!]
\centering
\begin{minipage}[h]{0.95\linewidth}
\centering
\includegraphics[width=0.99\linewidth]{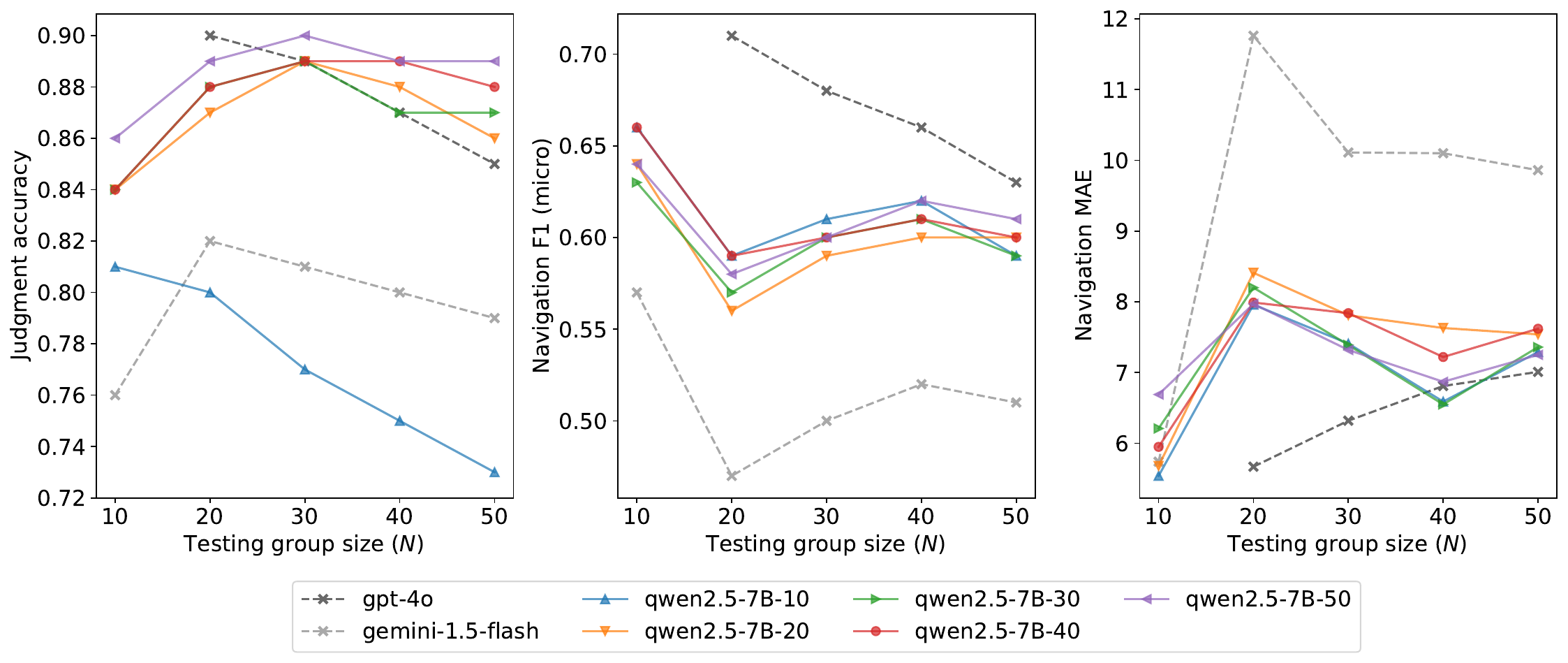}
\end{minipage}
\\
\begin{minipage}[h]{0.95\linewidth}
\centering
\includegraphics[width=0.99\linewidth]{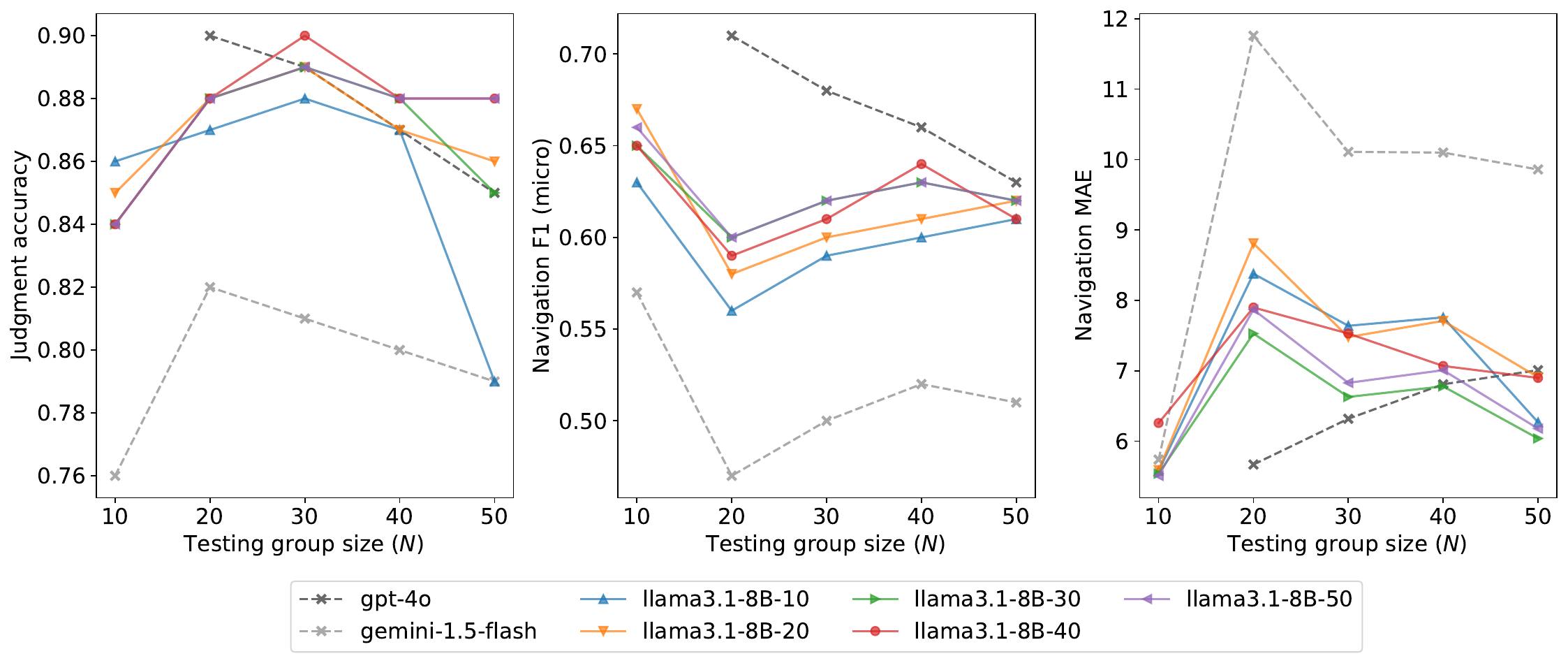}
\end{minipage}
\caption {Fine-tuned Qwen2.5-7B (first row) and Llama3.1-8B (second row) evaluation results with varying group size $N$.
The suffix of the model (i.e., \texttt{-10}) denotes the \textit{maximum} group size used in fine-tuning (details in \S\ref{subsec:dataset_development}). 
}
\label{fig:diff_number_train}
\end{figure*}

We also explore the use of public LLMs fine-tuned with transcripts associated with $N$ question-answer pairs (denoted by a suffix following the model name such as \texttt{Llama3.2-1B-10}, detailed in \Cref{tb:main_results} and \Cref{fig:diff_number_train}).
However, in our pilot studies, we observe that directly applying a fixed group size $N$ in all training data causes various problems, including frequently generating exact $N$ responses irrespective of the instruction, and exhibiting excessive repetition in responses.
To mitigate these issues, instead of fixing the group size $N$, we introduce variability in the training instances by sampling a random number \( K \in [5, N] \) and using $K$ question-answer pairs to construct each training instance.
For example, in our fine-tuning setup, \texttt{Llama3.2-1B-10} means that each transcript in the training set is paired with \textit{up to} 10 question-answer pairs, while the actual number is determined by randomness. Such stochastic approach enhances model robustness and reduces undesirable results in the generated responses.



\begin{table*}
\resizebox{0.995\textwidth}{!}{
\centering
\begin{tabular}{l|rrrrr|rrrrr|rrrrr||rrrrr}
\toprule
& \multicolumn{5}{c|}{\textbf{Judgment Accuracy} ($\uparrow$)} & \multicolumn{5}{c|}{\textbf{Navigation F1} ($\uparrow$)} & \multicolumn{5}{c||}{\textbf{Navigation MAE} ($\downarrow$)} & \multicolumn{5}{c}{\textbf{JSON Decode Error Rate ($\downarrow$)}} \\
 Group size $N$ & 10 & 20 & 30 & 40 & 50 & 10 & 20 & 30 & 40 & 50 & 10 & 20 & 30 & 40 & 50 & 10 & 20 & 30 & 40 & 50 \\\midrule
gpt-4o & / & \textbf{0.90} & \textbf{0.89} & \textbf{0.87} & \textbf{0.85} & / & \textbf{0.71} & \textbf{0.68} & \textbf{0.66} & \textbf{0.63} & / & \textbf{5.67} & \textbf{6.32} & 6.81 & 7.01 & 0 & 0 & 0 & 0 & 0 \\
gpt-4o-mini & 0.78 & 0.81 & 0.82 & 0.79 & 0.79 & 0.58 & 0.48 & 0.49 & 0.48 & 0.46 & 6.59 & 10.13 & 9.76 & 9.90 & 10.42 & 0 & 0 & 0 & 0 & 0 \\
gemini-1.5-flash & 0.76 & 0.82 & 0.81 & 0.80 & 0.79 & 0.57 & 0.47 & 0.50 & 0.52 & 0.51 & 5.74 & 11.76 & 10.11 & 10.10 & 9.86 & 0 & 0 & 0.01 & 0 & 0.03 \\\midrule
llama3.2-1B-inst & 0.58 & 0.55 & 0.55 & 0.53 & 0.48 & 0.19 & 0.16 & 0.13 & 0.13 & 0.11 & 13.25 & 14.70 & 14.98 & 15.84 & 16.90 & 0.48 & 0.58 & 0.59 & 0.62 & 0.73 \\
llama3.1-8B-inst & 0.65 & 0.61 & 0.59 & 0.55 & 0.53 & 0.29 & 0.20 & 0.21 & 0.20 & 0.19 & 9.40 & 11.52 & 12.17 & 13.12 & 13.61 & 0.31 & 0.39 & 0.39 & 0.38 & 0.49 \\
qwen2.5-7B-inst & 0.68 & 0.60 & 0.58 & 0.57 & 0.52 & 0.35 & 0.29 & 0.30 & 0.33 & 0.32 & 9.74 & 11.42 & 11.00 & 9.64 & 10.46 & 0.12 & 0.23 & 0.28 & 0.25 & 0.43 \\\midrule

llama3.2-1B-10 & 0.68 & 0.69 & 0.70 & 0.68 & 0.67 & 0.47 & 0.39 & 0.43 & 0.45 & 0.44 & 8.84 & 11.87 & 11.45 & 10.84 & 10.78 & 0 & 0 & 0 & 0 & 0.04 \\
llama3.1-8B-10 & \textbf{0.86} & 0.87 & 0.88 & 0.87 & 0.79 & 0.63 & 0.56 & 0.59 & 0.60 & 0.61 & 5.58 & 8.38 & 7.64 & 7.76 & \textbf{6.37} & 0 & 0 & 0 & 0 & 0.12 \\
qwen2.5-7B-10 & 0.81 & 0.80 & 0.77 & 0.75 & 0.73 & \textbf{0.66} & 0.59 & 0.61 & 0.62 & 0.59 & \textbf{5.54} & 7.96 & 7.41 & \textbf{6.59} & 7.29 & 0 & 0 & 0.01 & 0.03 & 0.01 \\
\bottomrule
\end{tabular}
}
\caption{Model evaluation results when varying group size $N$. 
The \texttt{-10} suffix of fine-tuned Llama and Qwen models denotes up to 10 question-answer pairs are paired with each transcript in the fine-tuning training set.
Since we use the generated responses from {GPT-4o} with $N$=10 as the standard reference, the corresponding cells are marked with ``/''. Detailed explanations for the above two settings are provided in \S\ref{subsec:dataset_development}.
}
\label{tb:main_results}
\end{table*}

\subsection{Experimental Settings}

\subsubsection{Models}

In this study, we evaluate both proprietary and public models for comprehensive benchmarking.

\paragraph{Proprietary Models.} 

We experiment with three popular commercial models: {GPT-4o} \cite{openai2024gpt4ocard}, {GPT-4o-mini} \cite{openai2024gpt4ocard}, and {Gemini-1.5-flash}.\footnote{\url{https://ai.google.dev/gemini-api/docs/models/gemini}}
{GPT-4o} is largely perceived as the best LLM for processing text and other modalities \cite{shahriar2024puttinggpt4oswordcomprehensive}, while {GPT-4o-mini} and {Gemini-1.5-flash} are the smaller and cheaper options better suited for a production environment.      
\paragraph{Llama-3 Models.}

Llama-3 \cite{grattafiori2024llama} represents the most recent iteration in the Llama series of open-source LLMs and is widely considered as one of the most advanced public models available. 
Due to the constraints of our industrial model serving environment, we only experiment with two size variations: \texttt{Llama3.2-1B} and \texttt{Llama3.1-8B}. 

For assessing the model's zero-shot capabilities, we utilize the instruction-tuned versions: \texttt{Llama3.2-1B-inst} and \texttt{Llama3.1-8B-inst}. Conversely, the untuned versions are employed for fine-tuning experiments. 
Model weights are accessed from Hugging Face \cite{wolf-etal-2020-transformers}.

\paragraph{Qwen-2.5 Models.}

Alibaba's Qwen-2.5 \cite{qwen2025qwen25technicalreport} is another family of advanced public LLMs, which is engineered to handle a diverse range of tasks with significant advancements in long-text generation and reasoning.  
In this study, we focus on the \texttt{Qwen2.5-7B} version to strike a balance between model performance and scalability. 
Similar to our setup for Llama-3 models, we select the instruction-tuned version, \texttt{Qwen2.5-7B-inst}, for zero-shot evaluations, while the untuned version is reserved for fine-tuning experiments.

\subsubsection{Metrics}

\paragraph{Judgment Accuracy.}

For the \judgment portion of the model responses, since it contains only a binary ``yes'' or ``no'' judgment, we directly use accuracy to evaluate these responses.


\paragraph{Navigation F1 and Mean Absolute Error.}

As detailed in \S\ref{subsec:our_method}, the second part of the model responses is $\mathsf{Navigation}$.
Since it is a multi-class classification task (i.e., assigning an index from $M$ utterances), we employ an F1 score to measure the model's ability to correctly identify each utterance. 

Since information can be spread over multiple conversation turns and several adjacent utterances may be relevant to the topic of a question, we incorporate Mean Absolute Error (MAE) as the additional metric to assess Navigation relevancy.
Specifically, our Navigation MAE is defined as:
\begin{align}
\text{Navigation MAE} = \frac{1}{N} \sum_{i=1}^{N} \left| y_i - \hat{y}_i \right|,
\nonumber
\end{align}
where \( N \) is the questions' group size, \( y_i \) and \( \hat{y}_i \) are the true and predicted navigation indices for the \( i \)-th prediction, respectively.


\subsubsection{Implementation Details}

Our subsequent experiments are conducted on a single node with 8 Nvidia A100 GPUs. 
For all our fine-tuning jobs, we utilize BFloat-16 \cite{bf16} precision and apply a learning rate of 3e-5 with 3 epochs.

\subsection{Results}

We vary the group size $N$ for generation and compare the performance of different models on the corresponding test sets.
Our experimental results are reported in \Cref{tb:main_results}.
Among all the proprietary models, {GPT-4o} achieves the best scores across all metrics, while {Gemini-1.5-flash} ranks the lowest.
We observe a decline in performance across all models as the group size $N$ increases, indicating that these LLMs struggle with handling a larger number of queries simultaneously.

Although the instruct versions of the public models cannot deliver equivalent performance as their proprietary counterparts, we find a significant performance increase in those public LLMs after fine-tuning with up to 10 grouped questions, achieving results comparable to — and in some cases surpassing — {Gemini-1.5-flash} and {GPT-4o-mini}. A more detailed analysis is provided in \S\ref{subsec:fine_tuning_analysis}.

\section{Analysis}

\subsection{Fine-Tuning Improves Model Performance}
\label{subsec:fine_tuning_analysis}

As shown in \Cref{tb:main_results}, we observe a significant performance increase in small public LLMs after fine-tuning with up to 10 grouped questions (using training data detailed in \S\ref{subsec:dataset_development}). We further explore and compare model performance after more granular fine-tuning with up to $N$ grouped questions, where \( N \in \{10, 20, 30, 40, 50\} \), as detailed in \S\ref{subsec:dataset_development}. 

Our experimental results are illustrated in \Cref{fig:diff_number_train}. It is evident that fine-tuned public LLMs show stronger capabilities in both \judgment and \navigation with more grouped questions included in the training process. For \judgment accuracy, GPT-4o maintains dominant performance with $N\leq20$; however, fine-tuned {Qwen2.5-7B} and {Llama3.1-8B} both surpass {GPT-4o} with $N\geq30$. This demonstrates the strong potential of smaller public LLMs.
For the \navigation F1 and MAE, {GPT-4o} still presents the best overall performance, but the gap between public LLMs and {GPT-4o} narrows noticeably after fine-tuning.

\subsection{JSON Decode Error}

In our task, answers to multiple questions are generated in a single model output. To enhance the parseability of the output in a production environment, we reinforce a JSON format on our responses (discussed in \S\ref{subsec:our_method}).
Additionally, we report the error rate of JSON decoding in \Cref{tb:main_results} to show the percentage of generated responses that contain any types of JSON decoding issue (key errors, quotation mark mismatches, etc). 

While such errors are hardly observed in proprietary LLMs, the instruction-tuned versions of both Llama3 and Qwen2.5 exhibit significant deficiencies in generating correct and desired JSON output structures. However, both these two public models show a substantial reduction in error rates across all group size after fine-tuning with only up to 10 grouped questions.

\section{Conclusion}
In this study, we evaluate the ability of LLMs to answer multiple questions based on the long conversational context within a single prompt. By adapting batch prompting, we efficiently address the challenge of processing multiple queries with conversational contexts. Through extensive experiments, we observe that while proprietary models like GPT-4o excel in the overall performance, fine-tuned smaller public models like Llama3.1-8B and Qwen2.5-7B can achieve comparable or even superior answer accuracy when more questions are grouped. Our findings demonstrate the viability of fine-tuning smaller models to enhance their capability in processing multiple questions, and underscore the potential for deploying cost-effective models without significant loss of accuracy.

\section*{Limitations}

Our study has the following limitations.
First, due to several practical constraints, we did not benchmark strong reasoning models emerged recently, such as DeepSeek R1 \cite{deepseekai2025deepseekr1incentivizingreasoningcapability}, QwQ \cite{qwq-32b-preview} and OpenAI o1 \cite{openai2024openaio1card}, which have demonstrated superior performance in public benchmarks.\footnote{\url{https://lmarena.ai/}}
Second, due to the nature of our specific QA use case, our experiments focused solely on Yes/No type questions. While our findings may still apply to a broader range of questions, future studies should validate these results using more types of questions.

\bibliography{custom}

\appendix
\section{Appendix}

\subsection{Example Questions}
\label{sec:appendix_example_questions}
We provide some example Yes/No type questions that can be useful in our customer support domain:
\begin{itemize}
    \item[-] Did the agent introduce themselves?
    \item[-] Was the account information verified?
    \item[-] Was the order confirmed?
    \item[-] Did the agent ask if further assistance was required?
    \item[-] Did the agent confirm the customer’s preferred communication channel?
\end{itemize}

\subsection{Prompt Template}
\label{sec:appendix_A}
Our prompt template is as follows:\\

{\fontfamily{qcr}\selectfont{\normalsize \small     
Here is a conversation transcript: \\
\begin{center}
    Utterance 1: \\
    Utterance 2:  \\
    ... \\
    Utterance M: \\
\end{center}

Answer the following questions based on the above conversation transcript: \\
\begin{center}
    Question 1: \\
    Question 2: \\
    Question 3: \\
    ... \\
    Question N: \\
\end{center}
The result should be in the JSON format, use the index of each question as the key, while the value should be an array consists of two parts: 
the first part is a string starting with a "Yes" or "No" answer to the question 
followed by justification. For the second part, return the index of one utterance
from the transcript that can best support your justification,
for "No" answers just use "NA" as the second part. Do not provide any judgment on conversation quality. An example of the output format:
\begin{verbatim}
{
    "Q1": ["Yes, the agent verified customer's 
    information at the start of the call", "5"], 
    "Q2": ["No, the agent did not send a copy 
    to the customer.", "NA"]
}
\end{verbatim}
}}

\end{document}